\documentclass[11pt]{article}

\usepackage[final]{acl}

\usepackage{times}
\usepackage{latexsym}

\usepackage[T1]{fontenc}

\usepackage[utf8]{inputenc}

\usepackage{microtype}

\usepackage{inconsolata}

\usepackage{graphicx}

\usepackage{amsmath}
\usepackage{subcaption}
\usepackage{booktabs}
\usepackage{hyperref} 
\usepackage{xurl}
\title{Fractional Rotation, Full Potential?\\Investigating Performance and Convergence of Partial RoPE}

\author{
 \textbf{Mohammad Aflah Khan\textsuperscript{1}},
 \textbf{Krishna P. Gummadi\textsuperscript{1}},
 \textbf{Manish Gupta\textsuperscript{2}},
 \textbf{Abhilasha Ravichander\textsuperscript{1}}
\\
 \textsuperscript{1}Max Planck Institute for Software Systems,
 \textsuperscript{2}Microsoft, Hyderabad
\\
 \small{
   \textbf{Correspondence:} \href{mailto:afkhan@mpi-sws.org}{afkhan@mpi-sws.org}
 }
}

\begin{document}
\maketitle
\begin{abstract}
Rotary Positional Embedding (RoPE) is widely used in transformers to encode relative positional information, yet the effect of applying RoPE to only a subset of hidden dimensions remains underexplored. This design choice can substantially reduce memory usage, especially at long context lengths. We show that partial RoPE achieves up to 10$\times$ memory savings over standard RoPE cache while maintaining comparable final loss. We systematically study its impact on training dynamics and convergence across architectures, model sizes, sequence lengths, and datasets. Our results reveal that: (1) applying RoPE to only a small fraction of dimensions (around 10\%) matches the convergence of full RoPE; (2) these trends remain consistent across settings; and (3) NoPE (No Positional Encoding) can produce loss spikes, which are mitigated by minimal RoPE usage or QK-Norm application. Overall, our findings highlight partial RoPE as an effective trade-off between efficiency and stability.
\end{abstract}

\section{Introduction}

Transformers use positional encodings to capture the order of tokens within a sequence, since the other parts of the model themselves are permutation-invariant. Rotary Positional Embedding (RoPE) \citep{su2023roformerenhancedtransformerrotary} has emerged as a leading method for encoding relative positions directly in the query-key interactions in self-attention \citep{vaswani2017attention} computations. RoPE is favored in modern decoder-only language models for its strong empirical performance and ability to generalize to sequences longer than those seen during training. It is widely adopted across architectures, either as the sole positional encoding or alongside alternatives such as No Positional Encoding (NoPE) \citep{kazemnejad2023impactpositionalencodinglength}.

\begin{figure}[!t]
\centering
\includegraphics[width=\linewidth]{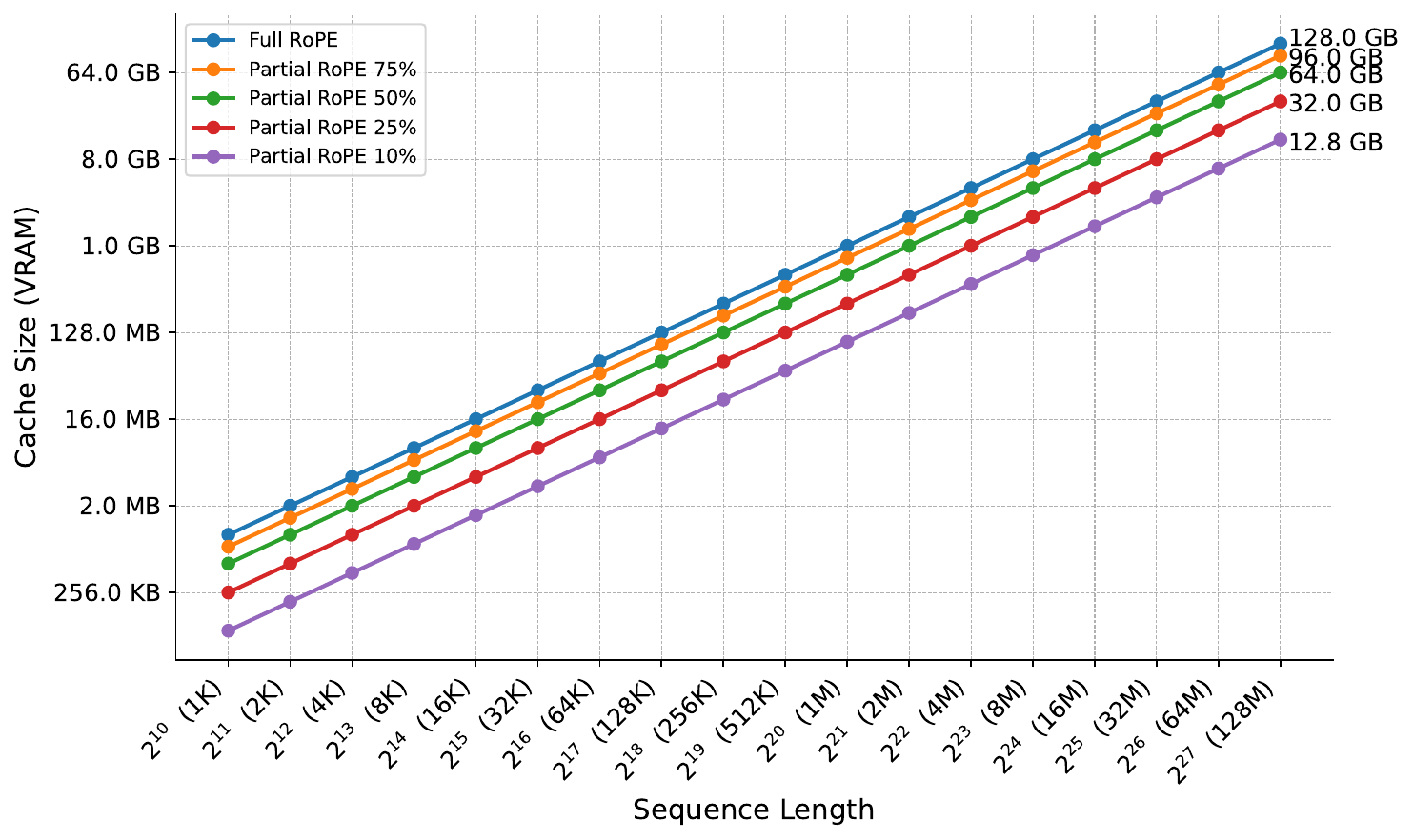}
\caption{Estimated RoPE cache memory usage as a function of sequence length. Applying RoPE to only a fraction of hidden dimensions substantially reduces cache size, yielding significant savings at long context lengths and in resource-constrained settings such as edge devices. More details are described in Appendix~\ref{app:rope_cache}.}
\label{fig:RoPE_Cache_VRAM_Size_vs_Sequence_Length}

\end{figure}

Despite the widespread adoption of RoPE, a fundamental design choice remains largely unexplored. What is the minimum fraction of hidden dimensions within each attention head that needs to undergo the rotary transformation? This parameter varies considerably across major model families, highlighting a lack of consensus on best practices. Early implementations in models like GPT-J \citep{gptj} and GPT-NeoX \citep{gptneox}, and later Pythia \citep{biderman2023pythia}, adopted a partial approach, applying RoPE to only 25\% of the dimensions. In contrast, the LLaMA series \citep{llama1,llama2,llama3,llama4} and most of the Qwen series \citep{qwen1,qwen2,qwen2_5,qwen3} applied the transformation to all dimensions. Interestingly, Qwen3-Next model moved to a 25\% application, a change claimed to improve extrapolation to longer sequences \citep{qwen2025qwen3Next}. Other models have explored intermediate values; Nemotron-4-340B \citep{nemotron340b} \& GLM-4.5 \citep{5team2025glm45agenticreasoningcoding} use 50\% while Phi-2 \citep{phi2} uses 40\%.


This wide variation in implementation highlights a gap in the literature, as no prior work has systematically studied how partial RoPE affects convergence, training dynamics, or efficiency. Unlike other architectural choices, partial RoPE is inconsistently applied, rarely ablated, and not well supported by most frameworks. Importantly, its memory implications become a significant design consideration primarily when training models with extremely long sequence lengths; at shorter or moderately long sequence lengths, RoPE-related cache memory is generally not a prominent constraint. In extreme long-context settings, however, applying RoPE to only a small fraction of dimensions (e.g. 10\%) can reduce cache memory by an order of magnitude (Fig.~\ref{fig:RoPE_Cache_VRAM_Size_vs_Sequence_Length}). A systematic study of this parameter can therefore improve our understanding of its effects and enable more memory-efficient long-context designs. To support reproducibility and further research, we release code to reproduce our setup\footnote{\url{www.github.com/aflah02/Partial_RoPE_Analysis}}, trained models\footnote{\url{hf.co/collections/aflah/partial-rope-analysis}} and loss trajectories.\footnote{\url{www.wandb.ai/aflah/rope-pct-exps/overview}}

\section{Related Work}


\textbf{RoPE and Partial RoPE.} RoPE~\citep{su2023roformerenhancedtransformerrotary} is widely used in transformer models because of its simplicity, strong length generalization, and compatibility with diverse attention mechanisms. Prior work has explored multiple aspects of RoPE design and integration. Several studies, as well as frontier models, omit RoPE in selected layers or combine it with sliding-window attention to improve efficiency while preserving performance \citep{kazemnejad2023impactpositionalencodinglength, commanda, llama4}. Other work has examined how the rotary base and scaling strategies influence convergence and long-context extrapolation \citep{men2024baseropeboundscontext, yang2025ropenopeagainnew}. Additional studies analyze the underlying mechanics of RoPE \citep{barbero2025roundroundgomakes}, explore guarantees on RoPE's distinguishing ability between tokens \citep{du2026ropedistinguishespositionstokens} and highlight precision-related issues that arise in long-context settings \citep{wang2024precision}. Despite these efforts, the specific question of how many dimensions of the hidden state should undergo rotation has received little attention. To our knowledge, the only discussion on partial RoPE application appears in \citet{gptneox}, which suggested, based on small-scale experiments,\footnote{On inspecting the public logs we were able to infer the authors used the GPT2-Small architecture (124M parameters) and trained on 82M tokens, in contrast we test multiple architectures at the 1B/8B scale and over 100B tokens} that a 25\% application offered a good trade-off between performance and efficiency. These findings have influenced design choices in multiple subsequent models. 

\textbf{Long Context Models.} There has been sustained momentum toward training models with ever-larger context windows. Notable examples include Google’s Gemini 1.5 family \citep{gemini_1_5} and Meta’s Llama 4 Scout \cite{llama4}, both of which report context windows of up to 10 million tokens, as well as preliminary work by Magic demonstrating 100 million token contexts \citep{magic}. Highly used frontier models also now boast context windows of a million tokens such as Opus 4.6 \citep{anthropic2026claudeopus46} and GPT-5.5 \citep{openai2026gpt55models}. At these extreme scales, previously negligible implementation details become critical. For instance, RoPE cache, which typically consumes minimal VRAM, can require substantial chunks of memory usage due to linear scaling with sequence length. This challenge is compounded in multi-GPU setups, where either replicating the cache on each device consumes redundant memory, or sharding it introduces communication overhead.

\begin{figure*}[ht]
    \centering
    \begin{subfigure}{0.48\linewidth}
        \centering
        \includegraphics[width=\linewidth]{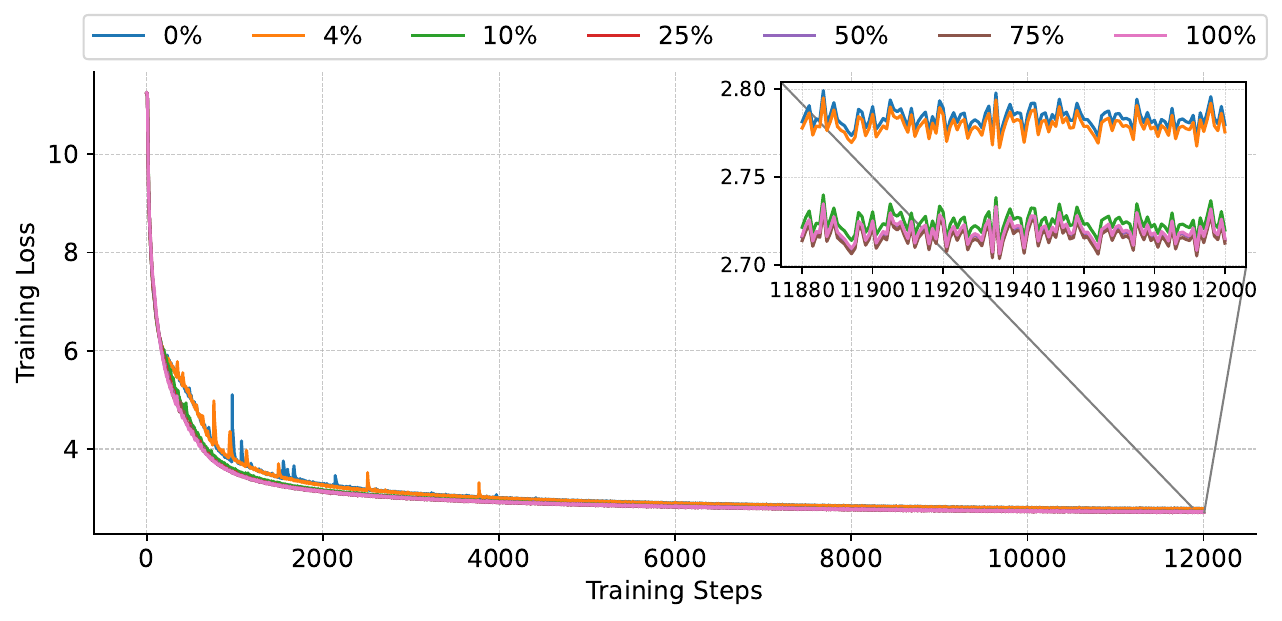}
        \caption{Default Sequence length (2048)}
        \label{fig:llama_3_2_1b_fw_loss_vs_steps_rope_pct}
    \end{subfigure}
    \hfill
    \begin{subfigure}{0.48\linewidth}
        \centering
        \includegraphics[width=\linewidth]{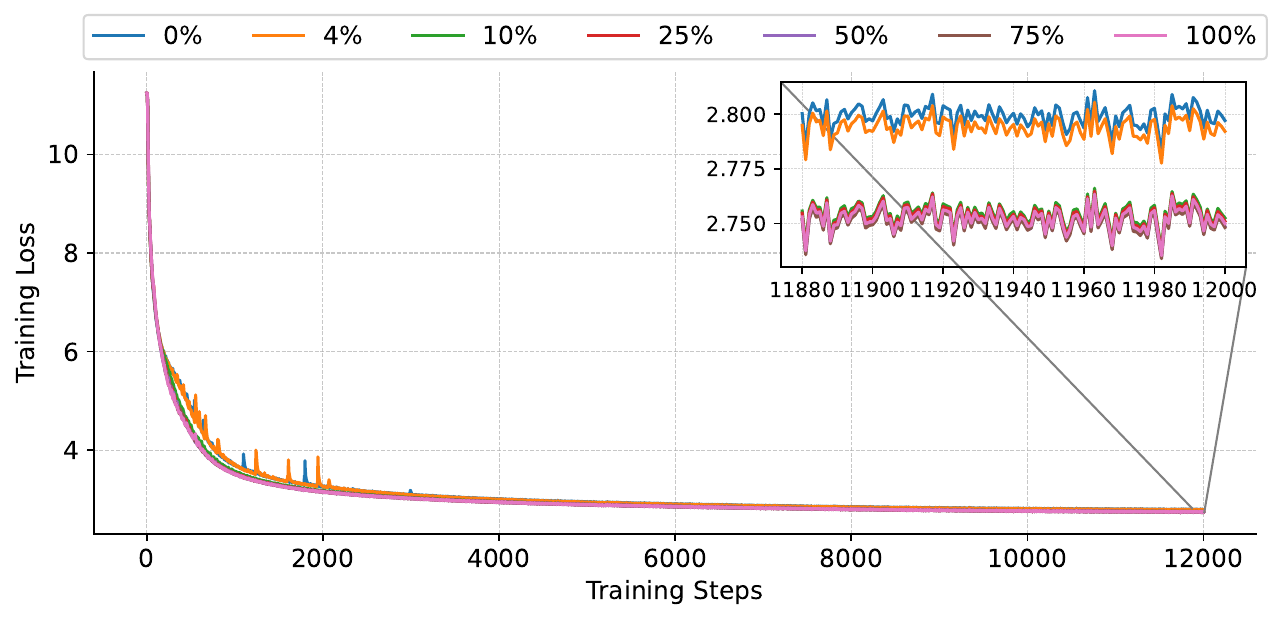}
        \caption{Sequence length 1024}
        \label{fig:llama_3_2_1b_fw_loss_vs_steps_rope_pct_sl_1024}
    \end{subfigure}


    \begin{subfigure}{0.48\linewidth}
        \centering
        \includegraphics[width=\linewidth]{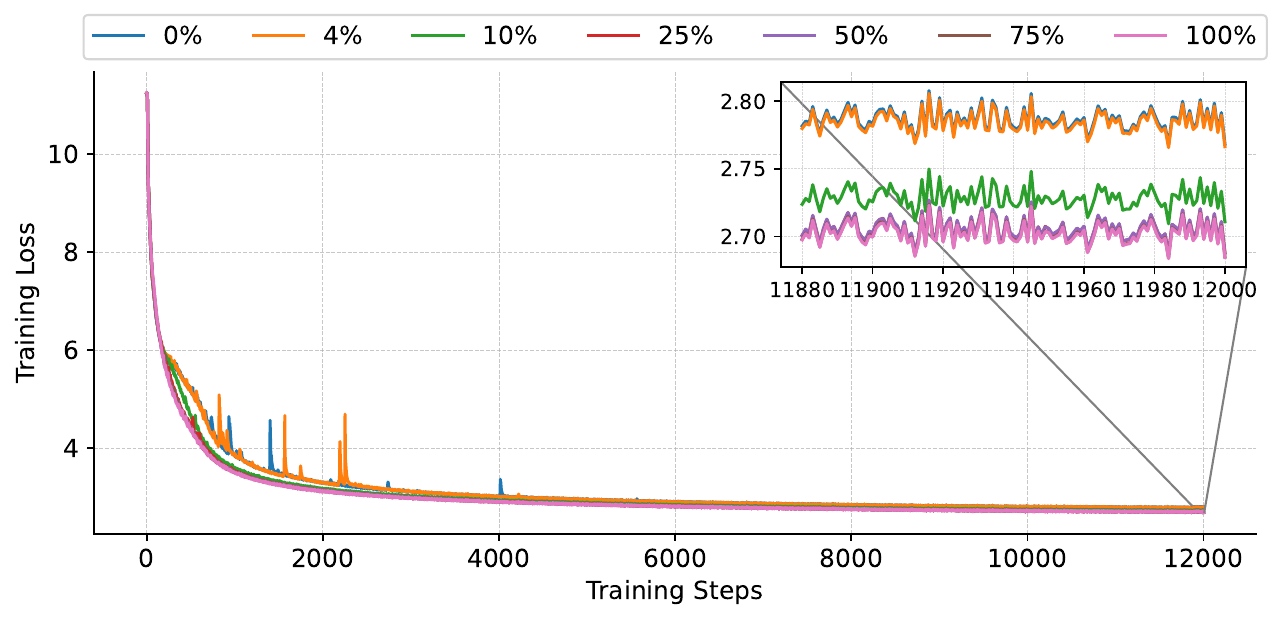}
        \caption{Sequence length 4096}
        \label{fig:llama_3_2_1b_fw_loss_vs_steps_rope_pct_sl_4096}
    \end{subfigure}
    \hfill
    \begin{subfigure}{0.48\linewidth}
        \centering
        \includegraphics[width=\linewidth]{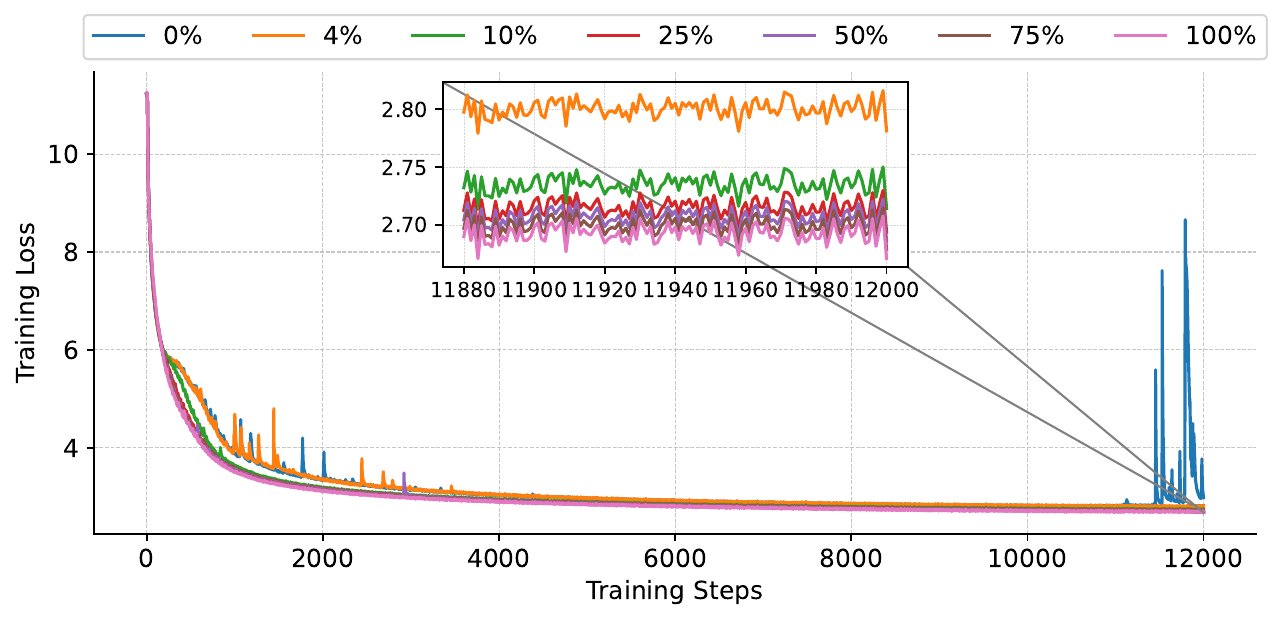}
        \caption{Sequence length 8192}
        \label{fig:llama_3_2_1b_fw_loss_vs_steps_rope_pct_sl_8192}
    \end{subfigure}

    \caption{Training loss trajectories on the FineWeb dataset for sequential models with varying RoPE configurations.}
    \label{fig:rope_pct_loss_fineweb}
\end{figure*}

\section{Experimental Setup}



We pretrain several models from scratch to examine how the fraction of each attention head’s hidden state receiving rotary positional embeddings affects model performance. Specifically, we apply RoPE to the first $K$\% of dimensions within each head and evaluate fractions 0\% (NoPE), 10\%, 25\%, 50\%, 75\%, and 100\% (full RoPE), to observe their effect on loss convergence. Additionally, we include experiments with the minimal feasible application, corresponding to just two channels per head, which represents approximately 1\% for Pythia-1B (head dimension 256) and 4\% for LLaMA-1B (head dimension 64). Further details are outlined in App.~\ref{app:rope_pct}.

We evaluate two transformer architectures: (1) a sequential attention design, implemented using Llama-3.2-1B and
Llama-3.1-8B \citep{llama3.2, llama3}, and (2) a parallel attention design, implemented using Pythia-1B \citep{biderman2023pythia}. The primary training is conducted on the FineWeb dataset \citep{fineweb}, with additional experiments on FineWeb-Edu \citep{finewebedu} to assess effect of dataset quality. We use the officially released 100B-token subsets and process all inputs with the Pythia tokenizer. 

In addition to loss, we evaluate models using EleutherAI’s LM Evaluation Harness \citep{eval-harness}. We adopt the same benchmarks used for the Pythia suite \citep{biderman2023pythia}, as they are well suited for non-instruction-tuned models, cover diverse task types, and remain informative at the 100B-token training scale \citep{wei2026hubble}. This set is further supplemented with PubMedQA \citep{jin2019pubmedqa}. Additional training and evaluation details are presented in Appendix~\ref{app:training_details}.

\section{Partial RoPE Analysis}


\textbf{RQ1: How does the fraction of hidden dimensions receiving RoPE influence model training dynamics?} As shown in Fig.~\ref{fig:llama_3_2_1b_fw_loss_vs_steps_rope_pct}, configurations using 10\% or more RoPE exhibit nearly identical convergence behavior. By the end of training, two distinct convergence groups emerge: models with NoPE or with RoPE applied to only two channels (4\%) converge to higher final losses, while those with 10\% or more RoPE achieve similar and lower final losses. This indicates that applying RoPE to even a modest fraction of hidden dimensions is sufficient to match the convergence of full RoPE.


\begin{figure*}[ht]
    \centering
    \begin{subfigure}{0.48\linewidth}
        \centering
        \includegraphics[width=\linewidth]{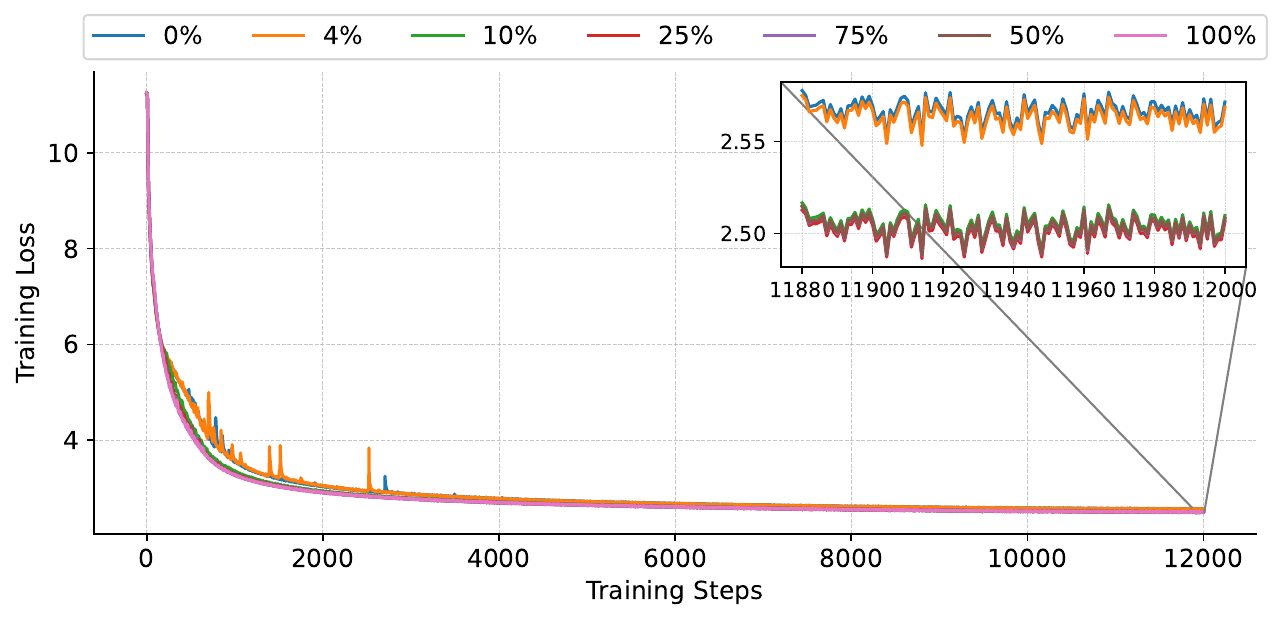}
        \caption{Sequential attention model.}
        \label{fig:llama_3_2_1b_fw_edu_loss_vs_steps_rope_pct}
    \end{subfigure}
    \hfill
    \begin{subfigure}{0.48\linewidth}
        \centering
        \includegraphics[width=\linewidth]{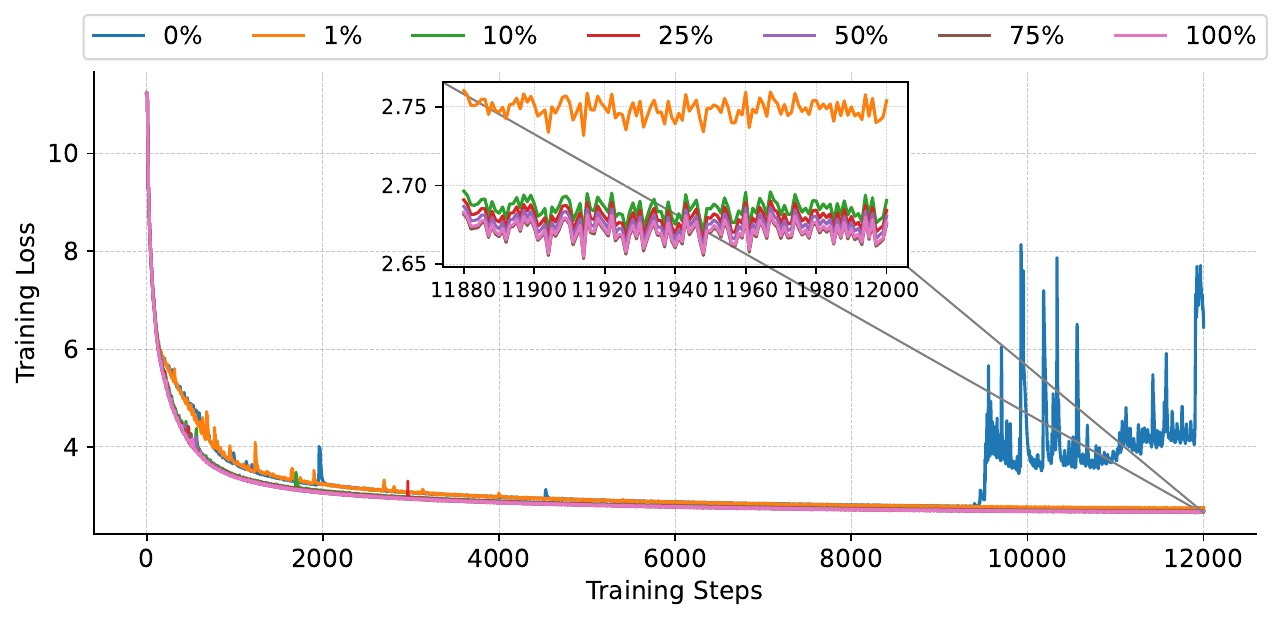}
        \caption{Parallel attention model.}
        \label{fig:pythia_1b_fw_edu_loss_vs_steps_rope_pct}
    \end{subfigure}

    \caption{Training loss trajectories on the FineWeb-Edu dataset comparing sequential and parallel attention architectures under varying Partial RoPE configurations.}
    \label{fig:rope_pct_loss_fineweb_edu}
\end{figure*}

\textbf{RQ2: How does the quality of pre-training data affect the optimal partial RoPE configuration?} To explore this, we repeated the experiments using FineWeb-Edu, a higher quality dataset. FineWeb-Edu is derived from FineWeb by applying a classifier that filters for content with higher educational quality. The impact of data quality is evident in the higher final loss values across runs (see Figs.~\ref{fig:llama_3_2_1b_fw_loss_vs_steps_rope_pct} and~\ref{fig:llama_3_2_1b_fw_edu_loss_vs_steps_rope_pct}). Nonetheless, similar convergence patterns are observed across both datasets.




\textbf{RQ3: How does the training sequence length influence the optimal Partial RoPE configuration?}
To evaluate the effect of sequence length on Partial RoPE behavior, we repeat the experiments from RQ1 using sequence lengths of 1024, 4096, and 8192 tokens, which correspond to commonly used pretraining context window sizes. As shown in Fig.~\ref{fig:llama_3_2_1b_fw_loss_vs_steps_rope_pct_sl_1024},~\ref{fig:llama_3_2_1b_fw_loss_vs_steps_rope_pct_sl_4096}, and~\ref{fig:llama_3_2_1b_fw_loss_vs_steps_rope_pct_sl_8192}, the models exhibit similar convergence bands across configurations, indicating that the observed trends are largely consistent across different sequence lengths.

We observe a single notable exception in which a loss spike appears for the NoPE run at a sequence length of 8192; strategies to mitigate this behavior are discussed later. Additionally, as the sequence length increases, the 10\% run begins to diverge slightly from the 25\% and higher settings. However, this separation is smaller than the differences observed for the NoPE and 4\% runs.


\begin{figure}
    \centering
    \includegraphics[width=\linewidth]{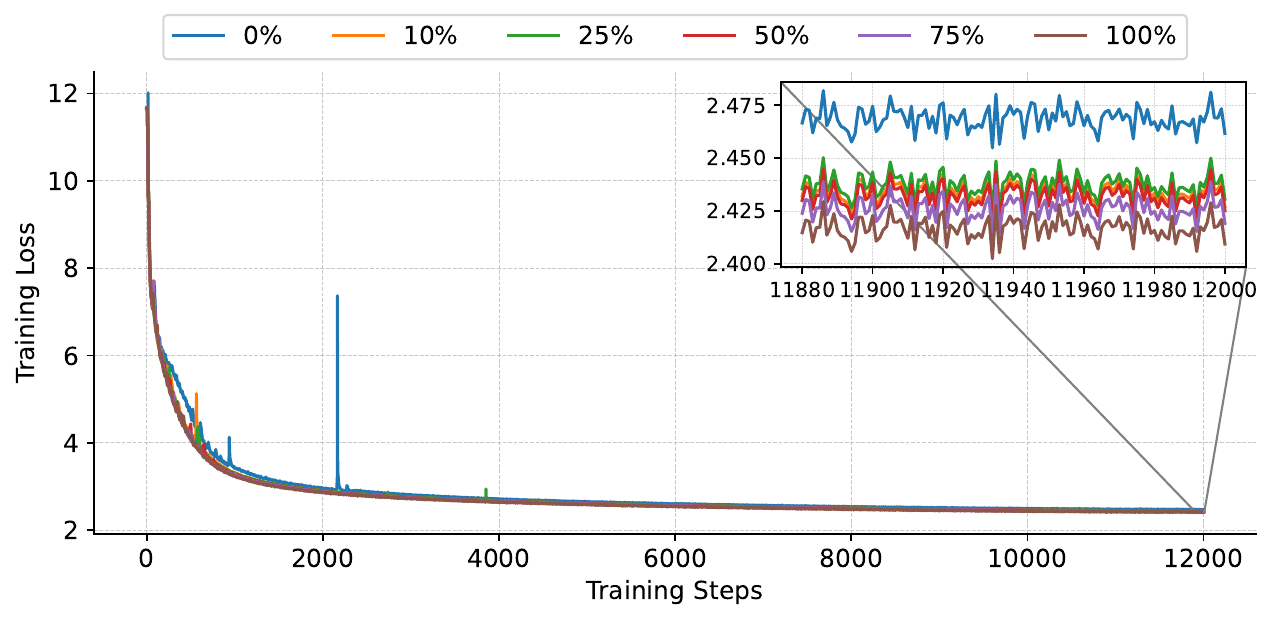}
    \caption{Training loss trajectories for the 8B model with varying Partial RoPE configurations.}
    \label{fig:llama_8b}
\end{figure}


\begin{figure*}[h]
    \centering
    \includegraphics[width=0.8\textwidth]{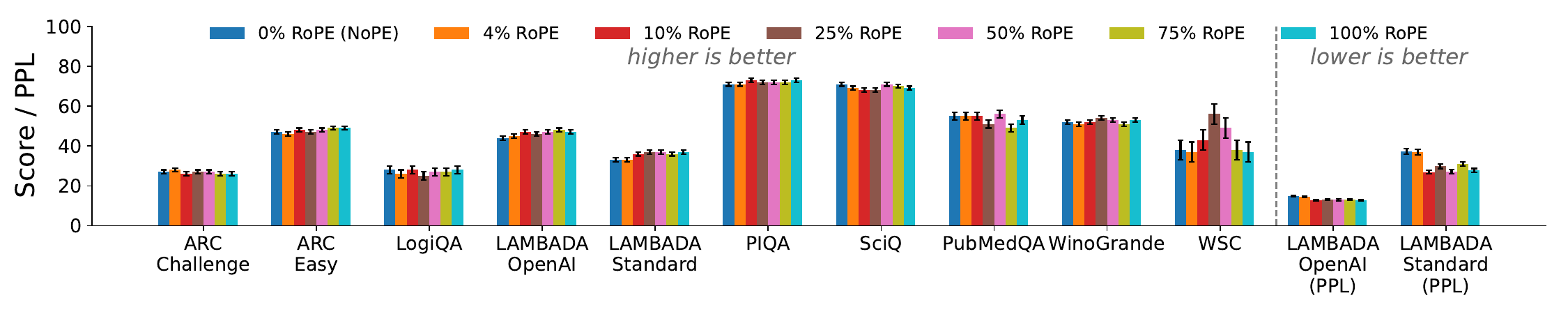} 
    \caption{Benchmark performance across tasks for the Sequential 1B model trained on FineWeb (Seq. Len. 2048)}
    \label{fig:eval_plot}
\end{figure*}

\textbf{RQ4: How consistent are the effects of partial RoPE across different transformer block designs (sequential vs. parallel attention)?} To investigate this, we trained models following the Pythia-1B architecture with parallel transformer blocks and observed two notable trends. The NoPE configuration (0\%) fails to converge (Fig.~\ref{fig:pythia_1b_fw_edu_loss_vs_steps_rope_pct}), a phenomenon we analyze further later. Excluding the NoPE run, two distinct convergence bands emerge, similar to those seen in sequential attention models, indicating that the trends hold across transformer block designs.

\textbf{RQ5: How do the effects of partial RoPE change with model scale?} To study scaling, we train a Llama-3-8B-like model on 100B FineWeb tokens. We again observe distinct convergence bands: NoPE forms a separate higher-loss cluster, while RoPE variants cluster together (Fig.~\ref{fig:llama_8b}). Relative to the 1B model at 2048 sequence length, RoPE configurations in the 8B model are slightly more dispersed, but overall trends remain consistent.


\textbf{RQ6: Do benchmark evaluation results corroborate the loss-based analysis?}
Across 9 out of 10 benchmarks, all RoPE variants exhibit broadly similar multiple-choice question (MCQ) performance, with WSC as the only exception, where no single configuration consistently outperforms the others. In contrast, LAMBADA perplexity drops substantially when transitioning from $<10\%$ to $\geq 10\%$ RoPE, while remaining relatively stable across higher-fraction variants (Fig.~\ref{fig:eval_plot}). Additional results are provided in Table~\ref{tab:pythia_general_eval_zeroshot} and ~\ref{tab:lambada_eval}. This is inline with the observations of \citet{heineman2025signalnoiseframeworkreducing}, where perplexity was found to offer a stronger signal as well as our findings of distinct loss bands.


\textbf{Takeaways.} Our experiments reveal clear trends for partial RoPE. Applying RoPE to even a small fraction of hidden dimensions (10\% or more) is enough to replicate the convergence behavior and final-loss performance of full RoPE. These patterns hold across datasets of varying quality, sequence lengths, and transformer block architectures, and remain consistent as model size scales from 1B to 8B parameters, though larger models exhibit slightly more variability across RoPE configurations. Evaluations beyond loss largely support these results: RoPE variants perform similarly on most MCQ benchmarks, while perplexity analyses show that configurations with 10\% or more RoPE behave comparably and outperform lower-percentage variants. Overall, partial RoPE appears to be a promising optimization strategy.

\subsection*{Analyzing Loss Spikes  with NoPE}
\label{sec:nope_spike}


Having shown that partial RoPE attains comparable convergence, we next analyze the two unrecoverable loss spikes observed in the NoPE configurations. Such loss spikes were not reported by \newcite{kazemnejad2023impactpositionalencodinglength}, who primarily studied smaller synthetic tasks with substantially lower model and dataset scales and did not consider parallel architectures. In contrast, we observe from our experiments that these loss spikes emerged at larger training scales. They appeared after tens of billions of tokens in parallel architectures, and only under similarly large training budgets combined with long context lengths in sequential architectures. To identify the potential root causes and mitigate these spikes, we systematically investigate the following for the parallel attention model.


\begin{figure*}[ht]
    \centering
    \begin{subfigure}{0.48\linewidth}
        \centering
        \includegraphics[width=\linewidth]{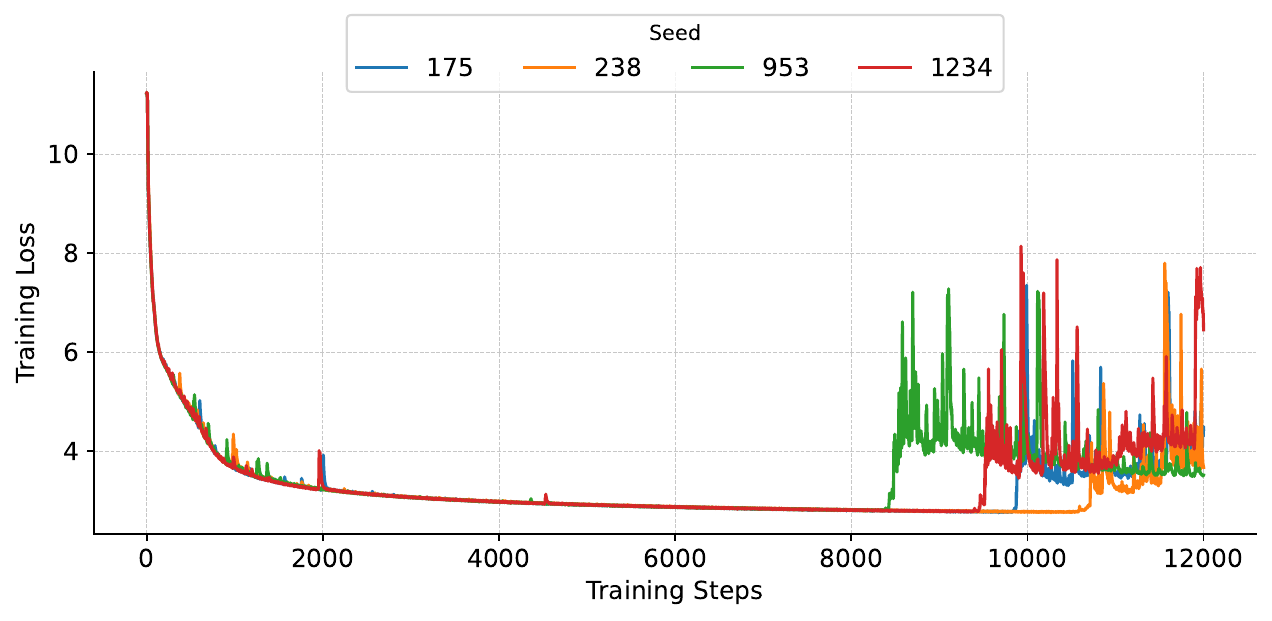}
        \caption{Different random seeds.}
        \label{fig:pythia_1b_fw_edu_loss_vs_steps_seeds}
    \end{subfigure}
    \hfill
    \begin{subfigure}{0.48\linewidth}
        \centering
        \includegraphics[width=\linewidth]{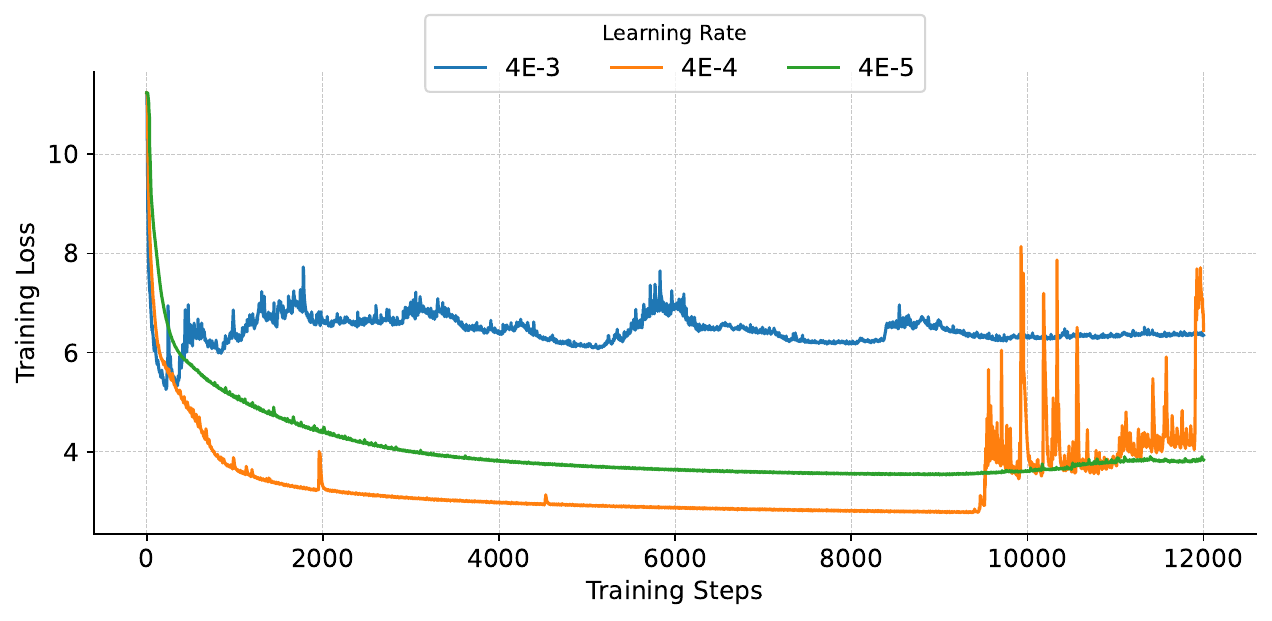}
        \caption{Different learning rates.}
        \label{fig:pythia_1b_fw_edu_loss_vs_steps_lrs}
    \end{subfigure}

    \caption{Training loss trajectories for parallel attention models trained on FineWeb-Edu under different experimental settings. The left panel varies random seeds, while the right panel varies learning rates.}
    \label{fig:pythia_1b_fw_edu_loss_vs_steps_combined}
\end{figure*}

\begin{figure*}[!ht]
    \centering

    \begin{subfigure}{0.48\linewidth}
        \centering
        \includegraphics[width=\linewidth]{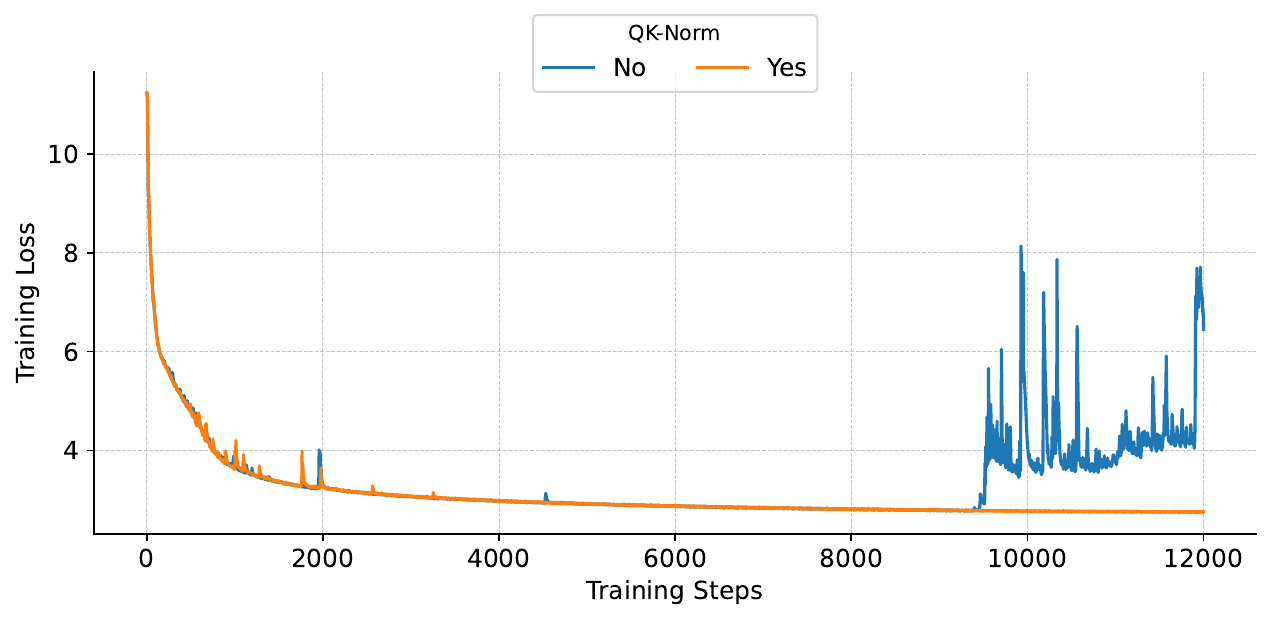}
        \caption{Pythia 1B (FineWeb-Edu): training loss trajectories with and without QK-Norm.}
        \label{fig:pythia_1b_fw_edu_loss_vs_steps_qk_norm}
    \end{subfigure}
    \hfill
    \begin{subfigure}{0.48\linewidth}
        \centering
        \includegraphics[width=\linewidth]{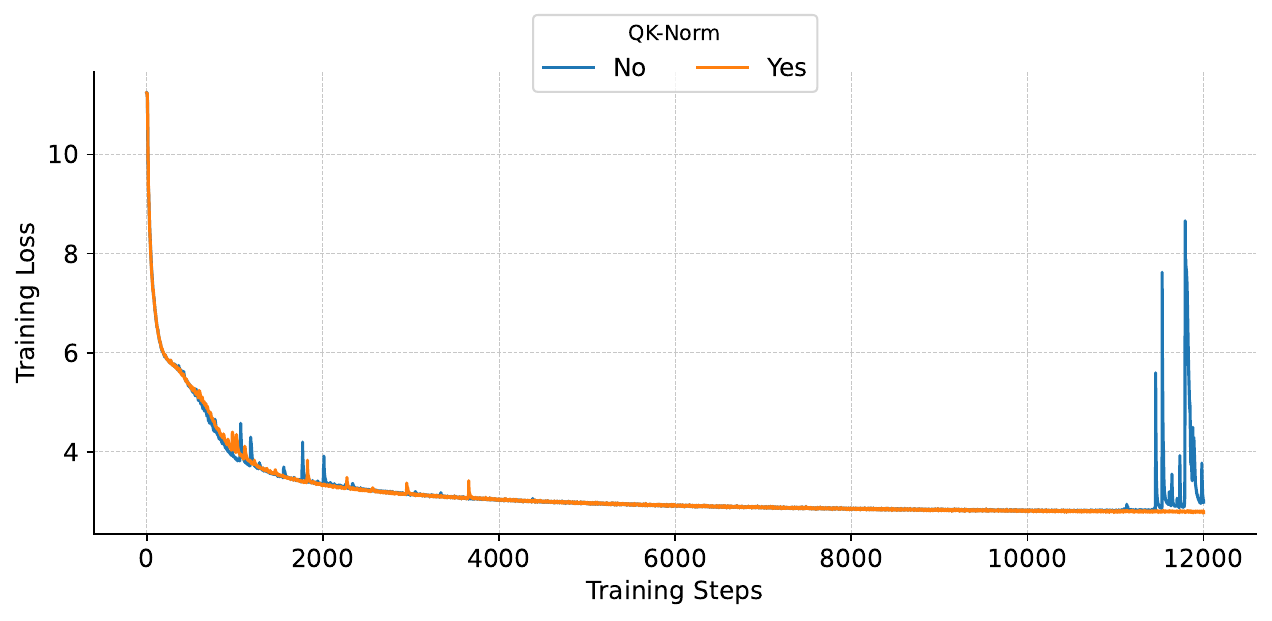}
        \caption{LLaMA 1B (FineWeb, seq len 8192): QK-Norm stabilizes training and removes loss spike seen without normalization.}
        \label{fig:llama_1b_seq_len_8192_fw_loss_vs_steps_qk_norm_last_channel}
    \end{subfigure}

    \caption{Comparison of training loss trajectories with and without QK-Norm across different 1B parameter models.}
    \label{fig:qk_norm_combined_loss_curves}
\end{figure*}

\textbf{Effect of Randomness:} We first considered that the spikes could result from an unlucky ordering of training data or model initialization. However, repeating the experiments with multiple random seeds (which change initialization and data order) consistently produced similar spikes (Fig.~\ref{fig:pythia_1b_fw_edu_loss_vs_steps_seeds}).


\textbf{Effect of Learning Rate (LR):} We next examine whether LR explains the  behavior. We tested learning rates an order of magnitude higher and lower than the base ($4\times10^{-3}$ \& $4\times10^{-5}$). With a lower LR, training converged to a higher loss, while a higher LR led to early spikes (Fig.~\ref{fig:pythia_1b_fw_edu_loss_vs_steps_lrs}).

\textbf{Effect of QK-Norm:} Finally, we applied QK-Norm \citep{qknorm_og}, a normalization technique known to stabilize training. More details about the implementation are outlined in Appendix~\ref{app:training_details}. With QK-Norm, the loss spikes disappear (Fig.~\ref{fig:pythia_1b_fw_edu_loss_vs_steps_qk_norm}), suggesting that the underlying cause may be excessively large or spiky gradients and parameter magnitudes \cite{qknorm_blog, olmo20252olmo2furious}, both of which this normalization mitigates. Following its application, the NoPE configuration converged to a higher loss band, consistent with patterns observed in other RQs. Similar trends are also observed in the sequential attention model as outlined in Fig.~\ref{fig:llama_1b_seq_len_8192_fw_loss_vs_steps_qk_norm_last_channel}. \textit{Based on this, we recommend that model trainers adopt QK-Norm as a precautionary measure against loss spikes or use partial RoPE to prevent the same.}

\section{Conclusion}



We presented a systematic empirical study of partial application of Rotary Positional Embeddings (RoPE) across hidden dimensions in large-scale transformer models. Our results show that even small RoPE fractions (as low as 10\% of dimensions depending on model architecture and sequence length) can match full RoPE in convergence and final loss, while extremely low fractions or NoPE configurations slow training and can introduce pronounced loss spikes. These patterns remain consistent across datasets of varying quality, model sizes, and both sequential and parallel transformer architectures. We also find that stabilization methods such as QK-Norm can reduce instability under NoPE.
Overall, partial RoPE offers a simple and effective way to reduce memory overhead (particularly for long-context and resource constrained settings) without sacrificing training stability or performance. Our findings provide practical guidance for transformer design and highlight partial RoPE as an underexplored but potentially impactful direction for efficient positional encoding.


\section*{Limitations}

As with any study involving pretraining, the design space is vast and each experiment incurs substantial computational cost. While it is infeasible to explore all possible combinations of architectures, model sizes, and datasets, we carefully select configurations that follow established best practices and evaluate them at scales consistent with prior work, allowing us to make claims we expect to generalize to larger settings.

We leave several directions for future work such as exploring the efficacy of the same with larger models, combining partial RoPE with NoPE for additional efficiency gains, investigating whether different layers or heads require different amounts of positional information,  studying scaling laws for partial RoPE, understanding the mechanism driving partial RoPE success and exploring the interaction of partial RoPE with length extrapolation methods due to the substantial computational resources and training time each of these investigations would require.



\bibliography{custom}

\appendix

\noindent {\Large \textbf{Overview of Appendix Sections}}
\begin{itemize}
  \item Appendix~\ref{app:rope_pct}: Rotary Dimension Allocation Details


  \item Appendix~\ref{app:training_details}: Training and Evaluation Details
  \begin{itemize}
    \item Training Details
    \item Evaluation Details
  \end{itemize}

  \item Appendix~\ref{app:rope_cache}: Computation of RoPE Cache Size

  \item Appendix~\ref{app:references}: References for Partial RoPE Usage

  \item Appendix~\ref{app:add_res}: Additional Results

  \item Appendix~\ref{app:llmusage}: LLM Usage

  \item Appendix~\ref{app:artifact}: Artifact Release/Usage
\end{itemize}
\section{Rotary Dimension Allocation Details}
\label{app:rope_pct}

We report the exact number of hidden dimensions to which RoPE is applied for each configuration. 
Because RoPE rotates hidden states in pairs, the number of rotated dimensions must be even. 
Accordingly, percentage-based specifications are rounded to the nearest valid even number of dimensions.

This adjustment only occurs for \textbf{Pythia}, which has a head dimension of 256. A nominal 10\% corresponds 
to 25.6 dimensions and would be rounded to 25 by the training framework. Since RoPE requires pairwise 
rotation, we instead use 26 dimensions (10.2\%) to ensure a valid pair count.

The resulting rotated dimensions for each model are listed in Tables~\ref{tab:rotation_dims}.

\begin{table*}[h]
\centering
\scriptsize
\begin{tabular}{lcccccccc}
\toprule
\textbf{Model (Head Dim)} & \textbf{0\%} & \textbf{1\%} & \textbf{4\%} & \textbf{10\% / 10.2\%} & \textbf{25\%} & \textbf{50\%} & \textbf{75\%} & \textbf{100\%} \\
\midrule
Pythia-1B (256) & 0 & 2 & -- & 26 & 64 & 128 & 192 & 256 \\
Llama-1B (64) & 0 & -- & 2 & 6 & 16 & 32 & 48 & 64 \\
Llama-8B (128) & 0 & -- & -- & 12 & 32 & 64 & 96 & 128 \\
\bottomrule
\end{tabular}
\caption{Number of rotated dimensions used for each RoPE percentage configuration. 
Values are adjusted to ensure an even number of dimensions since RoPE rotates hidden states pairwise.}
\label{tab:rotation_dims}
\end{table*}

\section{Training and Evaluation Details}
\label{app:training_details}

\subsection{Training Details}

All experiments are conducted on a cluster of 4 nodes, each equipped with 8 H200 GPUs. 

For most runs we use a per-GPU micro-batch size of 64 with 2 gradient accumulation steps, resulting in an effective global batch size of 4096 sequences, each containing 2048 tokens. For RQ3 experiments with a sequence length of 1024, we double the micro-batch size to 128 in order to keep the number of training steps consistent across runs. Conversely, for sequence lengths of 4096 and 8192, we halve and quarter the micro-batch size, respectively, to maintain the same training step count. For our 8B model runs we use a micro batch size of 4 with 32 gradient accumulation steps to attain the same effective global batch size of 4096 sequences.

Training is implemented using the GPT-NeoX framework \citep{gpt-neox-library}, selected for its strong support for partial RoPE and our existing familiarity with its ecosystem. Optimization is performed using AdamW \citep{adamw} with an initial learning rate of $4\times10^{-4}$ (unless otherwise specified), a 5\% warmup phase, and cosine learning rate decay to 10\% of the original learning rate \citep{cosineschedule}, without any dropout. All models are trained in mixed precision \citep{mixedprecisiontraining} following common conventions, where forward and backward computations are performed in BF16 while gradient accumulation and inter-GPU reductions are carried out in FP32. The training setup employs pure data parallelism across all GPUs, with no additional forms of model or pipeline parallelism. For 8B models, we follow \citet{wei2026hubble} by adding extra layers, resulting in a total of 36 layers compared to the 32 layers in Llama-3.1-8B. We also employ BF16 gradient accumulation with FP32 reductions to reduce memory usage.

We apply QK-Norm only over the hidden dimension of each attention head, rather than across both the attention heads and the head dimension. The latter is the default behavior in GPT-NeoX, which differs from the original QK-Norm formulation \citep{qknorm_og}. The original method also works with models that employ grouped query attention (GQA), such as the Llama-3.2-1B-based sequential models used in our experiments. 

For the 1B-parameter range, runs with parallel attention completed in 12–14 hours, while runs with sequential attention took 24–27 hours, with the longer sequence-length runs corresponding to the higher end of this range. Each 8B run took approximately 120 hours.

\subsection{Evaluation Details}

The benchmarks that we borrow from the Pythia Suite include ARC \citep{ai2_arc}, LogiQA \cite{liu2020logiqa}, LAMBADA \citep{lambadacite1, lambadacite2}, PIQA \citep{piqa}, SciQ \citep{SciQ}, WinoGrande \citep{ai2_winogrande}, and WSC \citep{wsc}.

For our MCQ evaluations, we primarily use byte-length normalized accuracy. However, for certain benchmarks, specifically Winogrande, PubMedQA, and WSC, LMEvalHarness only reports unnormalized accuracy, so we adopt that measure to remain consistent with prior work. Additionally, for LAMBADA, we consider perplexity alongside accuracy.

\begin{table*}[!ht]
    \small
    \begin{tabular}{ll}
        \toprule
        \textbf{Model} & \textbf{URL} \\
        \midrule
        EleutherAI/gpt-j-6b & \url{hf.co/EleutherAI/gpt-j-6b/blob/main/config.json#L21} \\
        EleutherAI/gpt-neox-20b & \url{hf.co/EleutherAI/gpt-neox-20b/blob/main/config.json#L19} \\
        EleutherAI/pythia-12b & \url{hf.co/EleutherAI/pythia-12b/blob/main/config.json#L17} \\
        microsoft/phi-2 & \url{hf.co/microsoft/phi-2/blob/main/config.json#L20} \\
        nvidia/Nemotron-4-340B-Base & \url{hf.co/nvidia/Nemotron-4-340B-Base/blob/main/model_config.yaml#L34} \\
        Qwen/Qwen3-Next-80B-A3B-Instruct & \url{hf.co/Qwen/Qwen3-Next-80B-A3B-Instruct/blob/main/config.json#L31} \\
        zai-org/GLM-4.5 & \url{hf.co/zai-org/GLM-4.5/blob/main/config.json#L16} \\
        \bottomrule
    \end{tabular}
    \caption{Examples of widely used open-weight/open-source models employing \textbf{Partial RoPE}, with references indicating this choice. Parameter names vary across models due to differences in training frameworks. Note: For Pythia, we show one family member, though all members use the same 25\% partial RoPE.}
    \label{tab:partial_rope_models}
\end{table*}

\section{Computation of RoPE Cache Size}
\label{app:rope_cache}

Rotary Positional Embeddings (RoPE) enhance transformer models by encoding positional information without learnable parameters. Instead of storing explicit embedding vectors for each position, RoPE applies a rotational transformation to query and key vectors on-the-fly. To optimize this process, the sine and cosine values required for these rotations are pre-computed and stored in a cache upon model initialization. 

The VRAM required to store the RoPE cache is a direct function of three key parameters: the maximum supported sequence length, the dimensionality of each attention head, and the numerical precision used for storage. The total size can be calculated using the following formula:

\begin{equation}
S_{cache} = L_{max} \times D_{head} \times P_{bytes}
\end{equation}

Where:
\begin{itemize}
\item \textit{S\textsubscript{cache}
 } is the total cache size in bytes.
\item \textit{L\textsubscript{max}
 } represents the maximum sequence length (or context window) the model is configured to handle.
\item \textit{D\textsubscript{head}
 } is the dimensionality of the vector for a single attention head.
\item \textit{P\textsubscript{bytes}
 } represents the number of bytes needed to store a single numerical value, determined by the chosen precision (e.g., 4 bytes for 32-bit floating point, FP32). We use FP32 in our calculations, as prior work has shown that lower precisions often lead to instability and divergence during long-context training, issues that are mitigated when using FP32 \citep{wang2024precision}.
\end{itemize}

This linear relationship demonstrates that the cache size scales directly with both the maximum sequence length and the head dimension, a critical consideration when designing models for long-context applications.

For Fig.~\ref{fig:RoPE_Cache_VRAM_Size_vs_Sequence_Length}, we use an attention head dimension of 256, which corresponds to the head dimensionality of Pythia-1B. The memory requirements would increase further for larger head dimensions (in practice most current models seem to stick to 128). Our calculations also exclude any effects of memory fragmentation and consider only the raw storage needed for the RoPE sine/cosine cache.

\section{References for Partial RoPE Usage}
\label{app:references}
Table~\ref{tab:partial_rope_models} lists various models that utilize partial RoPE, along with links referencing this configuration in their model settings.

\section{Additional Results}
\label{app:add_res}





Table~\ref{tab:pythia_general_eval_zeroshot} showcases the MCQ evaluation results for all models on all benchmarks while Table~\ref{tab:lambada_eval} showcases the perplexity evaluation results for all models on LAMBADA.

\section{LLM Usage}
\label{app:llmusage}
We use LLMs for help with grammatical corrections/writing of the paper and also as coding assistants. In both cases we independently verify the outputs.

\section{Artifact Release/Usage}
\label{app:artifact}
We publicly release all our artifacts (final models, optimizer states, config files, etc.) as linked in the main text.

Our usage of artifacts such as the pretraining dataset and frameworks is consistent with their intended usage.



\begin{table*}[ht]
\centering
\scriptsize
\setlength{\tabcolsep}{2.5pt} 

\begin{tabular}{lcccccccccc}
\toprule
\multicolumn{1}{c}{{\textbf{Config}}} &
  {\textbf{\begin{tabular}[c]{@{}c@{}}ARC\\ Challenge\end{tabular}}} &
  {\textbf{\begin{tabular}[c]{@{}c@{}}ARC\\ Easy\end{tabular}}} &
  {\textbf{LogiQA}} &
  {\textbf{\begin{tabular}[c]{@{}c@{}}LAMBADA \\ (OpenAI)\end{tabular}}} &
  {\textbf{\begin{tabular}[c]{@{}c@{}}LAMBADA \\ (Standard)\end{tabular}}} &
  {\textbf{PIQA}} &
  {\textbf{SciQ}} &
  {\textbf{PubMedQA}} &
  {\textbf{\begin{tabular}[c]{@{}c@{}}Wino\\ -Grande\end{tabular}}} &
  {\textbf{WSC}}\\

\midrule
\multicolumn{11}{c}{{\textbf{Llama-3.2-1B x FW-Edu x Seq. Len. 2048}}} \\
\midrule
  
{0\% RoPE (NoPE)} &
  {0.31 {\tiny (0.01)}} &
  {0.57 {\tiny (0.01)}} &
  {0.28 {\tiny (0.02)}} &
  {0.36 {\tiny (0.01)}} &
  {0.27 {\tiny (0.01)}} &
  {0.70 {\tiny (0.01)}} &
  {0.75 {\tiny (0.01)}} &
  {0.56 {\tiny (0.02)}} &
  {0.51 {\tiny (0.01)}} &
  {0.55 {\tiny (0.05)}}  \\
{4\% RoPE} &
  {0.31 {\tiny (0.01)}} &
  {0.58 {\tiny (0.01)}} &
  {0.28 {\tiny (0.02)}} &
  {0.36 {\tiny (0.01)}} &
  {0.27 {\tiny (0.01)}} &
  {0.69 {\tiny (0.01)}} &
  {0.75 {\tiny (0.01)}} &
  {0.51 {\tiny (0.02)}} &
  {0.52 {\tiny (0.01)}} &
  {0.38 {\tiny (0.05)}}  \\
{10\% RoPE} &
  {0.32 {\tiny (0.01)}} &
  {0.60 {\tiny (0.01)}} &
  {0.28 {\tiny (0.02)}} &
  {0.38 {\tiny (0.01)}} &
  {0.28 {\tiny (0.01)}} &
  {0.71 {\tiny (0.01)}} &
  {0.75 {\tiny (0.01)}} &
  {0.57 {\tiny (0.02)}} &
  {0.54 {\tiny (0.01)}} &
  {0.44 {\tiny (0.05)}}  \\
{25\% RoPE} &
  {0.31 {\tiny (0.01)}} &
  {0.58 {\tiny (0.01)}} &
  {0.26 {\tiny (0.02)}} &
  {0.38 {\tiny (0.01)}} &
  {0.29 {\tiny (0.01)}} &
  {0.70 {\tiny (0.01)}} &
  {0.75 {\tiny (0.01)}} &
  {0.52 {\tiny (0.02)}} &
  {0.53 {\tiny (0.01)}} &
  {0.54 {\tiny (0.05)}}  \\
{50\% RoPE} &
  {0.32 {\tiny (0.01)}} &
  {0.59 {\tiny (0.01)}} &
  {0.26 {\tiny (0.02)}} &
  {0.37 {\tiny (0.01)}} &
  {0.29 {\tiny (0.01)}} &
  {0.71 {\tiny (0.01)}} &
  {0.76 {\tiny (0.01)}} &
  {0.55 {\tiny (0.02)}} &
  {0.53 {\tiny (0.01)}} &
  {0.63 {\tiny (0.05)}}  \\
{75\% RoPE} &
  {0.33 {\tiny (0.01)}} &
  {0.61 {\tiny (0.01)}} &
  {0.27 {\tiny (0.02)}} &
  {0.39 {\tiny (0.01)}} &
  {0.27 {\tiny (0.01)}} &
  {0.71 {\tiny (0.01)}} &
  {0.75 {\tiny (0.01)}} &
  {0.48 {\tiny (0.02)}} &
  {0.52 {\tiny (0.01)}} &
  {0.62 {\tiny (0.05)}}  \\
{100\% RoPE} &
  {0.33 {\tiny (0.01)}} &
  {0.59 {\tiny (0.01)}} &
  {0.27 {\tiny (0.02)}} &
  {0.39 {\tiny (0.01)}} &
  {0.30 {\tiny (0.01)}} &
  {0.71 {\tiny (0.01)}} &
  {0.73 {\tiny (0.01)}} &
  {0.50 {\tiny (0.02)}} &
  {0.52 {\tiny (0.01)}} &
  {0.37 {\tiny (0.05)}}  \\

\midrule
\multicolumn{11}{c}{{\textbf{Llama-3.2-1B x FW x Seq. Len. 1024}}} \\
\midrule

{0\% RoPE (NoPE)} &
  {0.27 {\tiny (0.01)}} &
  {0.48 {\tiny (0.01)}} &
  {0.27 {\tiny (0.02)}} &
  {0.46 {\tiny (0.01)}} &
  {0.35 {\tiny (0.01)}} &
  {0.72 {\tiny (0.01)}} &
  {0.71 {\tiny (0.01)}} &
  {0.51 {\tiny (0.02)}} &
  {0.51 {\tiny (0.01)}} &
  {0.54 {\tiny (0.05)}}  \\
{4\% RoPE} &
  {0.27 {\tiny (0.01)}} &
  {0.50 {\tiny (0.01)}} &
  {0.28 {\tiny (0.02)}} &
  {0.46 {\tiny (0.01)}} &
  {0.33 {\tiny (0.01)}} &
  {0.71 {\tiny (0.01)}} &
  {0.70 {\tiny (0.01)}} &
  {0.50 {\tiny (0.02)}} &
  {0.53 {\tiny (0.01)}} &
  {0.48 {\tiny (0.05)}}  \\
{10\% RoPE} &
  {0.26 {\tiny (0.01)}} &
  {0.50 {\tiny (0.01)}} &
  {0.27 {\tiny (0.02)}} &
  {0.49 {\tiny (0.01)}} &
  {0.38 {\tiny (0.01)}} &
  {0.72 {\tiny (0.01)}} &
  {0.71 {\tiny (0.01)}} &
  {0.55 {\tiny (0.02)}} &
  {0.52 {\tiny (0.01)}} &
  {0.41 {\tiny (0.05)}}  \\
{25\% RoPE} &
  {0.26 {\tiny (0.01)}} &
  {0.49 {\tiny (0.01)}} &
  {0.28 {\tiny (0.02)}} &
  {0.48 {\tiny (0.01)}} &
  {0.38 {\tiny (0.01)}} &
  {0.72 {\tiny (0.01)}} &
  {0.72 {\tiny (0.01)}} &
  {0.55 {\tiny (0.02)}} &
  {0.55 {\tiny (0.01)}} &
  {0.37 {\tiny (0.05)}}  \\
{50\% RoPE} &
  {0.25 {\tiny (0.01)}} &
  {0.49 {\tiny (0.01)}} &
  {0.27 {\tiny (0.02)}} &
  {0.48 {\tiny (0.01)}} &
  {0.38 {\tiny (0.01)}} &
  {0.72 {\tiny (0.01)}} &
  {0.71 {\tiny (0.01)}} &
  {0.55 {\tiny (0.02)}} &
  {0.52 {\tiny (0.01)}} &
  {0.49 {\tiny (0.05)}}  \\
{75\% RoPE} &
  {0.26 {\tiny (0.01)}} &
  {0.49 {\tiny (0.01)}} &
  {0.27 {\tiny (0.02)}} &
  {0.47 {\tiny (0.01)}} &
  {0.38 {\tiny (0.01)}} &
  {0.73 {\tiny (0.01)}} &
  {0.71 {\tiny (0.01)}} &
  {0.56 {\tiny (0.02)}} &
  {0.53 {\tiny (0.01)}} &
  {0.62 {\tiny (0.05)}}  \\
{100\% RoPE} &
  {0.27 {\tiny (0.01)}} &
  {0.50 {\tiny (0.01)}} &
  {0.28 {\tiny (0.02)}} &
  {0.48 {\tiny (0.01)}} &
  {0.37 {\tiny (0.01)}} &
  {0.73 {\tiny (0.01)}} &
  {0.68 {\tiny (0.01)}} &
  {0.58 {\tiny (0.02)}} &
  {0.53 {\tiny (0.01)}} &
  {0.37 {\tiny (0.05)}}  \\

\midrule
\multicolumn{11}{c}{{\textbf{Llama-3.2-1B x FW x Seq. Len. 2048}}} \\
\midrule

{0\% RoPE (NoPE)} &
  {0.27 {\tiny (0.01)}} &
  {0.47 {\tiny (0.01)}} &
  {0.28 {\tiny (0.02)}} &
  {0.44 {\tiny (0.01)}} &
  {0.33 {\tiny (0.01)}} &
  {0.71 {\tiny (0.01)}} &
  {0.71 {\tiny (0.01)}} &
  {0.55 {\tiny (0.02)}} &
  {0.52 {\tiny (0.01)}} &
  {0.38 {\tiny (0.05)}}  \\
{4\% RoPE} &
  {0.28 {\tiny (0.01)}} &
  {0.46 {\tiny (0.01)}} &
  {0.26 {\tiny (0.02)}} &
  {0.45 {\tiny (0.01)}} &
  {0.33 {\tiny (0.01)}} &
  {0.71 {\tiny (0.01)}} &
  {0.69 {\tiny (0.01)}} &
  {0.55 {\tiny (0.02)}} &
  {0.51 {\tiny (0.01)}} &
  {0.37 {\tiny (0.05)}}  \\
{10\% RoPE} &
  {0.26 {\tiny (0.01)}} &
  {0.48 {\tiny (0.01)}} &
  {0.28 {\tiny (0.02)}} &
  {0.47 {\tiny (0.01)}} &
  {0.36 {\tiny (0.01)}} &
  {0.73 {\tiny (0.01)}} &
  {0.68 {\tiny (0.01)}} &
  {0.55 {\tiny (0.02)}} &
  {0.52 {\tiny (0.01)}} &
  {0.43 {\tiny (0.05)}}  \\
{25\% RoPE} &
  {0.27 {\tiny (0.01)}} &
  {0.47 {\tiny (0.01)}} &
  {0.25 {\tiny (0.02)}} &
  {0.46 {\tiny (0.01)}} &
  {0.37 {\tiny (0.01)}} &
  {0.72 {\tiny (0.01)}} &
  {0.68 {\tiny (0.01)}} &
  {0.51 {\tiny (0.02)}} &
  {0.54 {\tiny (0.01)}} &
  {0.56 {\tiny (0.05)}}  \\
{50\% RoPE} &
  {0.27 {\tiny (0.01)}} &
  {0.48 {\tiny (0.01)}} &
  {0.27 {\tiny (0.02)}} &
  {0.47 {\tiny (0.01)}} &
  {0.37 {\tiny (0.01)}} &
  {0.72 {\tiny (0.01)}} &
  {0.71 {\tiny (0.01)}} &
  {0.56 {\tiny (0.02)}} &
  {0.53 {\tiny (0.01)}} &
  {0.49 {\tiny (0.05)}}  \\
{75\% RoPE} &
  {0.26 {\tiny (0.01)}} &
  {0.49 {\tiny (0.01)}} &
  {0.27 {\tiny (0.02)}} &
  {0.48 {\tiny (0.01)}} &
  {0.36 {\tiny (0.01)}} &
  {0.72 {\tiny (0.01)}} &
  {0.70 {\tiny (0.01)}} &
  {0.49 {\tiny (0.02)}} &
  {0.51 {\tiny (0.01)}} &
  {0.38 {\tiny (0.05)}}  \\
{100\% RoPE} &
  {0.26 {\tiny (0.01)}} &
  {0.49 {\tiny (0.01)}} &
  {0.28 {\tiny (0.02)}} &
  {0.47 {\tiny (0.01)}} &
  {0.37 {\tiny (0.01)}} &
  {0.73 {\tiny (0.01)}} &
  {0.69 {\tiny (0.01)}} &
  {0.53 {\tiny (0.02)}} &
  {0.53 {\tiny (0.01)}} &
  {0.37 {\tiny (0.05)}}  \\

\midrule
\multicolumn{11}{c}{{\textbf{Llama-3.2-1B x FW x Seq. Len. 4096}}} \\
\midrule

{0\% RoPE (NoPE)} &
  {0.26 {\tiny (0.01)}} &
  {0.45 {\tiny (0.01)}} &
  {0.27 {\tiny (0.02)}} &
  {0.43 {\tiny (0.01)}} &
  {0.30 {\tiny (0.01)}} &
  {0.70 {\tiny (0.01)}} &
  {0.68 {\tiny (0.01)}} &
  {0.56 {\tiny (0.02)}} &
  {0.52 {\tiny (0.01)}} &
  {0.60 {\tiny (0.05)}}  \\
{4\% RoPE} &
  {0.25 {\tiny (0.01)}} &
  {0.46 {\tiny (0.01)}} &
  {0.26 {\tiny (0.02)}} &
  {0.42 {\tiny (0.01)}} &
  {0.30 {\tiny (0.01)}} &
  {0.70 {\tiny (0.01)}} &
  {0.69 {\tiny (0.01)}} &
  {0.56 {\tiny (0.02)}} &
  {0.52 {\tiny (0.01)}} &
  {0.37 {\tiny (0.05)}}  \\
{10\% RoPE} &
  {0.27 {\tiny (0.01)}} &
  {0.47 {\tiny (0.01)}} &
  {0.29 {\tiny (0.02)}} &
  {0.46 {\tiny (0.01)}} &
  {0.35 {\tiny (0.01)}} &
  {0.72 {\tiny (0.01)}} &
  {0.69 {\tiny (0.01)}} &
  {0.56 {\tiny (0.02)}} &
  {0.52 {\tiny (0.01)}} &
  {0.47 {\tiny (0.05)}}  \\
{25\% RoPE} &
  {0.26 {\tiny (0.01)}} &
  {0.47 {\tiny (0.01)}} &
  {0.27 {\tiny (0.02)}} &
  {0.47 {\tiny (0.01)}} &
  {0.34 {\tiny (0.01)}} &
  {0.72 {\tiny (0.01)}} &
  {0.69 {\tiny (0.01)}} &
  {0.46 {\tiny (0.02)}} &
  {0.53 {\tiny (0.01)}} &
  {0.42 {\tiny (0.05)}}  \\
{50\% RoPE} &
  {0.26 {\tiny (0.01)}} &
  {0.47 {\tiny (0.01)}} &
  {0.29 {\tiny (0.02)}} &
  {0.46 {\tiny (0.01)}} &
  {0.36 {\tiny (0.01)}} &
  {0.72 {\tiny (0.01)}} &
  {0.68 {\tiny (0.01)}} &
  {0.54 {\tiny (0.02)}} &
  {0.52 {\tiny (0.01)}} &
  {0.63 {\tiny (0.05)}}  \\
{75\% RoPE} &
  {0.27 {\tiny (0.01)}} &
  {0.48 {\tiny (0.01)}} &
  {0.27 {\tiny (0.02)}} &
  {0.46 {\tiny (0.01)}} &
  {0.36 {\tiny (0.01)}} &
  {0.71 {\tiny (0.01)}} &
  {0.73 {\tiny (0.01)}} &
  {0.56 {\tiny (0.02)}} &
  {0.53 {\tiny (0.01)}} &
  {0.64 {\tiny (0.05)}}  \\
{100\% RoPE} &
  {0.27 {\tiny (0.01)}} &
  {0.48 {\tiny (0.01)}} &
  {0.28 {\tiny (0.02)}} &
  {0.48 {\tiny (0.01)}} &
  {0.37 {\tiny (0.01)}} &
  {0.72 {\tiny (0.01)}} &
  {0.68 {\tiny (0.01)}} &
  {0.58 {\tiny (0.02)}} &
  {0.52 {\tiny (0.01)}} &
  {0.40 {\tiny (0.05)}}  \\

\midrule
\multicolumn{11}{c}{{\textbf{Llama-3.2-1B x FW x Seq. Len. 8192}}} \\
\midrule

{0\% RoPE (NoPE)} &
  {0.23 {\tiny (0.01)}} &
  {0.39 {\tiny (0.01)}} &
  {0.26 {\tiny (0.02)}} &
  {0.23 {\tiny (0.01)}} &
  {0.16 {\tiny (0.01)}} &
  {0.64 {\tiny (0.01)}} &
  {0.61 {\tiny (0.02)}} &
  {0.52 {\tiny (0.02)}} &
  {0.52 {\tiny (0.01)}} &
  {0.37 {\tiny (0.05)}}  \\
{0\% RoPE (NoPE) + QK-Norm} &
  {0.26 {\tiny (0.01)}} &
  {0.45 {\tiny (0.01)}} &
  {0.27 {\tiny (0.02)}} &
  {0.40 {\tiny (0.01)}} &
  {0.30 {\tiny (0.01)}} &
  {0.70 {\tiny (0.01)}} &
  {0.68 {\tiny (0.01)}} &
  {0.55 {\tiny (0.02)}} &
  {0.50 {\tiny (0.01)}} &
  {0.38 {\tiny (0.05)}}  \\
{4\% RoPE} &
  {0.25 {\tiny (0.01)}} &
  {0.45 {\tiny (0.01)}} &
  {0.26 {\tiny (0.02)}} &
  {0.41 {\tiny (0.01)}} &
  {0.30 {\tiny (0.01)}} &
  {0.71 {\tiny (0.01)}} &
  {0.68 {\tiny (0.01)}} &
  {0.55 {\tiny (0.02)}} &
  {0.53 {\tiny (0.01)}} &
  {0.66 {\tiny (0.05)}}  \\
{10\% RoPE} &
  {0.25 {\tiny (0.01)}} &
  {0.46 {\tiny (0.01)}} &
  {0.27 {\tiny (0.02)}} &
  {0.45 {\tiny (0.01)}} &
  {0.35 {\tiny (0.01)}} &
  {0.71 {\tiny (0.01)}} &
  {0.69 {\tiny (0.01)}} &
  {0.50 {\tiny (0.02)}} &
  {0.51 {\tiny (0.01)}} &
  {0.37 {\tiny (0.05)}}  \\
{25\% RoPE} &
  {0.26 {\tiny (0.01)}} &
  {0.47 {\tiny (0.01)}} &
  {0.26 {\tiny (0.02)}} &
  {0.45 {\tiny (0.01)}} &
  {0.34 {\tiny (0.01)}} &
  {0.71 {\tiny (0.01)}} &
  {0.71 {\tiny (0.01)}} &
  {0.55 {\tiny (0.02)}} &
  {0.52 {\tiny (0.01)}} &
  {0.37 {\tiny (0.05)}}  \\
{50\% RoPE} &
  {0.27 {\tiny (0.01)}} &
  {0.47 {\tiny (0.01)}} &
  {0.28 {\tiny (0.02)}} &
  {0.47 {\tiny (0.01)}} &
  {0.33 {\tiny (0.01)}} &
  {0.71 {\tiny (0.01)}} &
  {0.69 {\tiny (0.01)}} &
  {0.46 {\tiny (0.02)}} &
  {0.52 {\tiny (0.01)}} &
  {0.47 {\tiny (0.05)}}  \\
{75\% RoPE} &
  {0.25 {\tiny (0.01)}} &
  {0.46 {\tiny (0.01)}} &
  {0.29 {\tiny (0.02)}} &
  {0.47 {\tiny (0.01)}} &
  {0.37 {\tiny (0.01)}} &
  {0.71 {\tiny (0.01)}} &
  {0.69 {\tiny (0.01)}} &
  {0.55 {\tiny (0.02)}} &
  {0.53 {\tiny (0.01)}} &
  {0.38 {\tiny (0.05)}}  \\
{100\% RoPE} &
  {0.25 {\tiny (0.01)}} &
  {0.47 {\tiny (0.01)}} &
  {0.28 {\tiny (0.02)}} &
  {0.47 {\tiny (0.01)}} &
  {0.35 {\tiny (0.01)}} &
  {0.72 {\tiny (0.01)}} &
  {0.69 {\tiny (0.01)}} &
  {0.53 {\tiny (0.02)}} &
  {0.53 {\tiny (0.01)}} &
  {0.63 {\tiny (0.05)}}  \\
  
\midrule
\multicolumn{11}{c}{{\textbf{Pythia-1B x FW-Edu x Seq. Len. 2048}}} \\
\midrule

{0\% RoPE (NoPE)} &
  {0.24 {\tiny (0.01)}} &
  {0.30 {\tiny (0.01)}} &
  {0.24 {\tiny (0.02)}} &
  {0.00 {\tiny (0.00)}} &
  {0.00 {\tiny (0.00)}} &
  {0.53 {\tiny (0.01)}} &
  {0.28 {\tiny (0.01)}} &
  {0.34 {\tiny (0.02)}} &
  {0.50 {\tiny (0.01)}} &
  {0.63 {\tiny (0.05)}}  \\
{0\% RoPE (NoPE) + QK-Norm} &
  {0.28 {\tiny (0.01)}} &
  {0.52 {\tiny (0.01)}} &
  {0.29 {\tiny (0.02)}} &
  {0.28 {\tiny (0.01)}} &
  {0.22 {\tiny (0.01)}} &
  {0.67 {\tiny (0.01)}} &
  {0.71 {\tiny (0.01)}} &
  {0.53 {\tiny (0.02)}} &
  {0.52 {\tiny (0.01)}} &
  {0.37 {\tiny (0.05)}}  \\
{1\% RoPE} &
  {0.28 {\tiny (0.01)}} &
  {0.50 {\tiny (0.01)}} &
  {0.25 {\tiny (0.02)}} &
  {0.30 {\tiny (0.01)}} &
  {0.23 {\tiny (0.01)}} &
  {0.66 {\tiny (0.01)}} &
  {0.72 {\tiny (0.01)}} &
  {0.54 {\tiny (0.02)}} &
  {0.51 {\tiny (0.01)}} &
  {0.37 {\tiny (0.05)}}  \\
{10\% RoPE} &
  {0.29 {\tiny (0.01)}} &
  {0.53 {\tiny (0.01)}} &
  {0.27 {\tiny (0.02)}} &
  {0.34 {\tiny (0.01)}} &
  {0.24 {\tiny (0.01)}} &
  {0.68 {\tiny (0.01)}} &
  {0.73 {\tiny (0.01)}} &
  {0.48 {\tiny (0.02)}} &
  {0.51 {\tiny (0.01)}} &
  {0.37 {\tiny (0.05)}}  \\
{25\% RoPE} &
  {0.30 {\tiny (0.01)}} &
  {0.55 {\tiny (0.01)}} &
  {0.27 {\tiny (0.02)}} &
  {0.34 {\tiny (0.01)}} &
  {0.23 {\tiny (0.01)}} &
  {0.68 {\tiny (0.01)}} &
  {0.72 {\tiny (0.01)}} &
  {0.45 {\tiny (0.02)}} &
  {0.51 {\tiny (0.01)}} &
  {0.55 {\tiny (0.05)}}  \\
{50\% RoPE} &
  {0.29 {\tiny (0.01)}} &
  {0.54 {\tiny (0.01)}} &
  {0.27 {\tiny (0.02)}} &
  {0.33 {\tiny (0.01)}} &
  {0.23 {\tiny (0.01)}} &
  {0.67 {\tiny (0.01)}} &
  {0.73 {\tiny (0.01)}} &
  {0.48 {\tiny (0.02)}} &
  {0.51 {\tiny (0.01)}} &
  {0.41 {\tiny (0.05)}}  \\
{75\% RoPE} &
  {0.30 {\tiny (0.01)}} &
  {0.55 {\tiny (0.01)}} &
  {0.27 {\tiny (0.02)}} &
  {0.35 {\tiny (0.01)}} &
  {0.24 {\tiny (0.01)}} &
  {0.67 {\tiny (0.01)}} &
  {0.72 {\tiny (0.01)}} &
  {0.52 {\tiny (0.02)}} &
  {0.51 {\tiny (0.01)}} &
  {0.37 {\tiny (0.05)}}  \\
{100\% RoPE} &
  {0.28 {\tiny (0.01)}} &
  {0.54 {\tiny (0.01)}} &
  {0.27 {\tiny (0.02)}} &
  {0.34 {\tiny (0.01)}} &
  {0.24 {\tiny (0.01)}} &
  {0.69 {\tiny (0.01)}} &
  {0.72 {\tiny (0.01)}} &
  {0.55 {\tiny (0.02)}} &
  {0.51 {\tiny (0.01)}} &
  {0.37 {\tiny (0.05)}}  \\

\midrule
\multicolumn{11}{c}{{\textbf{Llama-3.1-8B x FW x Seq. Len. 2048}}} \\
\midrule

{0\% RoPE (NoPE)} &
  {0.30 {\tiny (0.01)}} &
  {0.57 {\tiny (0.01)}} &
  {0.28 {\tiny (0.02)}} &
  {0.58 {\tiny (0.01)}} &
  {0.49 {\tiny (0.01)}} &
  {0.77 {\tiny (0.01)}} &
  {0.80 {\tiny (0.01)}} &
  {0.60 {\tiny (0.02)}} &
  {0.57 {\tiny (0.01)}} &
  {0.42 {\tiny (0.05)}}  \\
{10\% RoPE} &
  {0.32 {\tiny (0.01)}} &
  {0.60 {\tiny (0.01)}} &
  {0.27 {\tiny (0.02)}} &
  {0.60 {\tiny (0.01)}} &
  {0.50 {\tiny (0.01)}} &
  {0.77 {\tiny (0.01)}} &
  {0.81 {\tiny (0.01)}} &
  {0.64 {\tiny (0.02)}} &
  {0.59 {\tiny (0.01)}} &
  {0.55 {\tiny (0.05)}}  \\
{25\% RoPE} &
  {0.33 {\tiny (0.01)}} &
  {0.58 {\tiny (0.01)}} &
  {0.27 {\tiny (0.02)}} &
  {0.59 {\tiny (0.01)}} &
  {0.52 {\tiny (0.01)}} &
  {0.77 {\tiny (0.01)}} &
  {0.82 {\tiny (0.01)}} &
  {0.60 {\tiny (0.02)}} &
  {0.59 {\tiny (0.01)}} &
  {0.59 {\tiny (0.05)}}  \\
{50\% RoPE} &
  {0.31 {\tiny (0.01)}} &
  {0.58 {\tiny (0.01)}} &
  {0.29 {\tiny (0.02)}} &
  {0.60 {\tiny (0.01)}} &
  {0.51 {\tiny (0.01)}} &
  {0.77 {\tiny (0.01)}} &
  {0.81 {\tiny (0.01)}} &
  {0.57 {\tiny (0.02)}} &
  {0.59 {\tiny (0.01)}} &
  {0.41 {\tiny (0.05)}}  \\
{75\% RoPE} &
  {0.33 {\tiny (0.01)}} &
  {0.58 {\tiny (0.01)}} &
  {0.29 {\tiny (0.02)}} &
  {0.60 {\tiny (0.01)}} &
  {0.52 {\tiny (0.01)}} &
  {0.77 {\tiny (0.01)}} &
  {0.79 {\tiny (0.01)}} &
  {0.59 {\tiny (0.02)}} &
  {0.58 {\tiny (0.01)}} &
  {0.44 {\tiny (0.05)}}  \\
{100\% RoPE} &
  {0.33 {\tiny (0.01)}} &
  {0.59 {\tiny (0.01)}} &
  {0.28 {\tiny (0.02)}} &
  {0.60 {\tiny (0.01)}} &
  {0.52 {\tiny (0.01)}} &
  {0.77 {\tiny (0.01)}} &
  {0.81 {\tiny (0.01)}} &
  {0.62 {\tiny (0.02)}} &
  {0.59 {\tiny (0.01)}} &
  {0.49 {\tiny (0.05)}}  \\

\bottomrule
\end{tabular}
\caption{\textbf{MCQ Benchmark Results.} We run evaluations through EleutherAI's Language Model Evaluation Harness \citep{eval-harness} under Zero-Shot setting.
}
\label{tab:pythia_general_eval_zeroshot}
\end{table*}

\begin{table*}
\centering
\scriptsize

\begin{tabular}{cccc}
\hline
\textbf{Config} & \textbf{LAMBADA (OpenAI)} & \textbf{LAMBADA (Standard)} & \textbf{LAMBADA} \\ \hline
\multicolumn{4}{c}{Llama-3.2-1B x FW-Edu x 2048}        \\ \hline
{0\% RoPE (NoPE)} &
  {26.72 {\scriptsize (0.96)}} &
  {69.87 {\scriptsize (2.90)}} &
  {48.30 {\scriptsize (11.00)}}  \\
{4\% RoPE} &
  {27.71 {\scriptsize (1.00)}} &
  {68.78 {\scriptsize (2.78)}} &
  {48.25 {\scriptsize (10.48)}}  \\
{10\% RoPE} &
  {23.25 {\scriptsize (0.82)}} &
  {64.50 {\scriptsize (2.63)}} &
  {43.87 {\scriptsize (10.50)}}  \\
{25\% RoPE} &
  {23.51 {\scriptsize (0.83)}} &
  {64.78 {\scriptsize (2.75)}} &
  {44.14 {\scriptsize (10.52)}}  \\
{50\% RoPE} &
  {23.89 {\scriptsize (0.85)}} &
  {64.87 {\scriptsize (2.75)}} &
  {44.38 {\scriptsize (10.45)}}  \\
{75\% RoPE} &
  {22.97 {\scriptsize (0.81)}} &
  {76.38 {\scriptsize (3.26)}} &
  {49.67 {\scriptsize (13.56)}}  \\
{100\% RoPE} &
  {22.99 {\scriptsize (0.82)}} &
  {57.82 {\scriptsize (2.35)}} &
  {40.41 {\scriptsize (8.88)}}  \\ \hline
\multicolumn{4}{c}{Llama-3.2-1B x FW x Seq. Len. 1024}  \\ \hline
{0\% RoPE (NoPE)} &
  {13.49 {\scriptsize (0.41)}} &
  {31.95 {\scriptsize (1.15)}} &
  {22.72 {\scriptsize (4.70)}}  \\
{4\% RoPE} &
  {13.41 {\scriptsize (0.41)}} &
  {32.38 {\scriptsize (1.17)}} &
  {22.90 {\scriptsize (4.82)}}  \\
{10\% RoPE} &
  {11.80 {\scriptsize (0.36)}} &
  {25.88 {\scriptsize (0.92)}} &
  {18.84 {\scriptsize (3.59)}}  \\
{25\% RoPE} &
  {12.35 {\scriptsize (0.37)}} &
  {27.11 {\scriptsize (0.94)}} &
  {19.73 {\scriptsize (3.76)}}  \\
{50\% RoPE} &
  {12.10 {\scriptsize (0.37)}} &
  {25.64 {\scriptsize (0.90)}} &
  {18.87 {\scriptsize (3.45)}}  \\
{75\% RoPE} &
  {12.35 {\scriptsize (0.37)}} &
  {24.07 {\scriptsize (0.83)}} &
  {18.21 {\scriptsize (3.00)}}  \\
{100\% RoPE} &
  {12.33 {\scriptsize (0.37)}} &
  {26.86 {\scriptsize (0.95)}} &
  {19.59 {\scriptsize (3.70)}}  \\ \hline
\multicolumn{4}{c}{Llama-3.2-1B x FW x Seq. Len. 2048}  \\ \hline
{0\% RoPE (NoPE)} &
  {14.90 {\scriptsize (0.46)}} &
  {37.15 {\scriptsize (1.38)}} &
  {26.03 {\scriptsize (5.66)}}  \\
{4\% RoPE} &
  {14.55 {\scriptsize (0.45)}} &
  {36.95 {\scriptsize (1.36)}} &
  {25.75 {\scriptsize (5.69)}}  \\
{10\% RoPE} &
  {12.86 {\scriptsize (0.39)}} &
  {26.91 {\scriptsize (0.94)}} &
  {19.89 {\scriptsize (3.59)}}  \\
{25\% RoPE} &
  {13.07 {\scriptsize (0.40)}} &
  {29.79 {\scriptsize (1.09)}} &
  {21.43 {\scriptsize (4.26)}}  \\
{50\% RoPE} &
  {12.96 {\scriptsize (0.40)}} &
  {27.11 {\scriptsize (0.97)}} &
  {20.03 {\scriptsize (3.62)}}  \\
{75\% RoPE} &
  {13.09 {\scriptsize (0.40)}} &
  {30.93 {\scriptsize (1.14)}} &
  {22.01 {\scriptsize (4.54)}}  \\
{100\% RoPE} &
  {12.78 {\scriptsize (0.39)}} &
  {27.73 {\scriptsize (0.98)}} &
  {20.26 {\scriptsize (3.81)}}  \\ \hline
\multicolumn{4}{c}{Llama-3.2-1B x FW x Seq. Len. 4096}  \\ \hline
{0\% RoPE (NoPE)} &
  {16.03 {\scriptsize (0.51)}} &
  {43.96 {\scriptsize (1.64)}} &
  {30.00 {\scriptsize (7.09)}}  \\
{4\% RoPE} &
  {16.61 {\scriptsize (0.53)}} &
  {43.40 {\scriptsize (1.63)}} &
  {30.01 {\scriptsize (6.80)}}  \\
{10\% RoPE} &
  {13.39 {\scriptsize (0.41)}} &
  {30.37 {\scriptsize (1.09)}} &
  {21.88 {\scriptsize (4.33)}}  \\
{25\% RoPE} &
  {13.48 {\scriptsize (0.42)}} &
  {33.71 {\scriptsize (1.24)}} &
  {23.60 {\scriptsize (5.14)}}  \\
{50\% RoPE} &
  {13.43 {\scriptsize (0.41)}} &
  {30.68 {\scriptsize (1.12)}} &
  {22.05 {\scriptsize (4.40)}}  \\
{75\% RoPE} &
  {13.22 {\scriptsize (0.40)}} &
  {31.09 {\scriptsize (1.14)}} &
  {22.16 {\scriptsize (4.55)}}  \\
{100\% RoPE} &
  {12.93 {\scriptsize (0.40)}} &
  {27.90 {\scriptsize (1.01)}} &
  {20.41 {\scriptsize (3.82)}}  \\ \hline
\multicolumn{4}{c}{Llama-3.2-1B x FW x Seq. Len. 8192}  \\ \hline
{0\% RoPE (NoPE)} &
  {79.69 {\scriptsize (3.37)}} &
  {396.00 {\scriptsize (20.03)}} &
  {237.85 {\scriptsize (80.38)}}  \\
{0\% RoPE (NoPE) + QK-Norm} &
  {19.09 {\scriptsize (0.62)}} &
  {47.90 {\scriptsize (1.85)}} &
  {33.49 {\scriptsize (7.33)}}  \\
{4\% RoPE} &
  {18.83 {\scriptsize (0.61)}} &
  {49.06 {\scriptsize (1.91)}} &
  {33.95 {\scriptsize (7.69)}}  \\
{10\% RoPE} &
  {14.39 {\scriptsize (0.45)}} &
  {32.69 {\scriptsize (1.20)}} &
  {23.54 {\scriptsize (4.66)}}  \\
{25\% RoPE} &
  {14.64 {\scriptsize (0.46)}} &
  {38.92 {\scriptsize (1.53)}} &
  {26.78 {\scriptsize (6.18)}}  \\
{50\% RoPE} &
  {14.15 {\scriptsize (0.45)}} &
  {40.83 {\scriptsize (1.57)}} &
  {27.49 {\scriptsize (6.77)}}  \\
{75\% RoPE} &
  {13.76 {\scriptsize (0.43)}} &
  {32.29 {\scriptsize (1.21)}} &
  {23.02 {\scriptsize (4.72)}}  \\
{100\% RoPE} &
  {13.55 {\scriptsize (0.42)}} &
  {34.64 {\scriptsize (1.31)}} &
  {24.10 {\scriptsize (5.36)}}  \\ \hline
\multicolumn{4}{c}{Pythia-1B x FW-Edu x Seq. Len. 2048} \\ \hline
{0\% RoPE (NoPE)} &
  {340933.20 {\scriptsize (25150.68)}} &
  {3717492.05 {\scriptsize (300190.33)}} &
  {2029212.62 {\scriptsize (870637.70)}}  \\
{0\% RoPE (NoPE) + QK-Norm} &
  {51.74 {\scriptsize (2.04)}} &
  {136.88 {\scriptsize (5.85)}} &
  {94.31 {\scriptsize (21.73)}}  \\
{1\% RoPE} &
  {43.13 {\scriptsize (1.69)}} &
  {144.46 {\scriptsize (6.33)}} &
  {93.79 {\scriptsize (25.75)}}  \\
{10\% RoPE} &
  {33.49 {\scriptsize (1.27)}} &
  {115.78 {\scriptsize (5.06)}} &
  {74.64 {\scriptsize (20.90)}}  \\
{25\% RoPE} &
  {35.18 {\scriptsize (1.34)}} &
  {140.81 {\scriptsize (6.30)}} &
  {88.00 {\scriptsize (26.80)}}  \\
{50\% RoPE} &
  {33.35 {\scriptsize (1.26)}} &
  {131.37 {\scriptsize (5.78)}} &
  {82.36 {\scriptsize (24.86)}}  \\
{75\% RoPE} &
  {32.52 {\scriptsize (1.23)}} &
  {114.03 {\scriptsize (4.93)}} &
  {73.27 {\scriptsize (20.69)}}  \\
{100\% RoPE} &
  {32.88 {\scriptsize (1.24)}} &
  {122.90 {\scriptsize (5.35)}} &
  {77.89 {\scriptsize (22.84)}}  \\ \hline
\multicolumn{4}{c}{Llama-3.1-8B x FW x Seq. Len. 2048}  \\ \hline
{0\% RoPE (NoPE)} &
  {6.80 {\scriptsize (0.18)}} &
  {11.01 {\scriptsize (0.32)}} &
  {8.91 {\scriptsize (1.09)}}  \\
{10\% RoPE} &
  {6.42 {\scriptsize (0.17)}} &
  {10.08 {\scriptsize (0.29)}} &
  {8.25 {\scriptsize (0.94)}}  \\
{25\% RoPE} &
  {6.37 {\scriptsize (0.16)}} &
  {9.66 {\scriptsize (0.27)}} &
  {8.01 {\scriptsize (0.85)}}  \\
{50\% RoPE} &
  {6.18 {\scriptsize (0.16)}} &
  {10.05 {\scriptsize (0.29)}} &
  {8.12 {\scriptsize (1.00)}}  \\
{75\% RoPE} &
  {6.07 {\scriptsize (0.15)}} &
  {9.34 {\scriptsize (0.26)}} &
  {7.71 {\scriptsize (0.85)}}  \\
{100\% RoPE} &
  {6.11 {\scriptsize (0.15)}} &
  {9.03 {\scriptsize (0.25)}} &
  {7.57 {\scriptsize (0.76)}}  \\ \hline

\end{tabular}
\caption{\textbf{Perplexity results for LAMBADA.} We run evaluations through EleutherAI's Language Model Evaluation Harness \citep{eval-harness} under Zero-Shot setting. 
}

\label{tab:lambada_eval}
\end{table*}

\end{document}